\documentclass[sigconf]{acmart}

\AtBeginDocument{%
  \providecommand\BibTeX{{%
    \normalfont B\kern-0.5em{\scshape i\kern-0.25em b}\kern-0.8em\TeX}}}

\setcopyright{acmcopyright}
\copyrightyear{2018}
\acmYear{2018}
\acmDOI{10.1145/1122445.1122456}

\acmConference[Woodstock '18]{Woodstock '18: ACM Symposium on Neural
  Gaze Detection}{June 03--05, 2018}{Woodstock, NY}
\acmBooktitle{Woodstock '18: ACM Symposium on Neural Gaze Detection,
  June 03--05, 2018, Woodstock, NY}
\acmPrice{15.00}
\acmISBN{978-1-4503-XXXX-X/18/06}

\usepackage[T1]{fontenc}
\usepackage[utf8]{inputenc}
\usepackage{amsmath}
\usepackage{subfigure}
\usepackage{longtable}
\usepackage{booktabs} 
\usepackage{multirow}
\usepackage{multicol}
\usepackage{color}

\usepackage{balance}

\copyrightyear{2022} 
\acmYear{2022} 
\setcopyright{acmcopyright}\acmConference[WWW '22]{Proceedings of the ACM Web Conference 2022}{April 25--29, 2022}{Virtual Event, Lyon, France}
\acmBooktitle{Proceedings of the ACM Web Conference 2022 (WWW '22), April 25--29, 2022, Virtual Event, Lyon, France}
\acmPrice{15.00}
\acmDOI{10.1145/3485447.3511998}
\acmISBN{978-1-4503-9096-5/22/04}

\begin{document}

\title{KnowPrompt:  Knowledge-aware Prompt-tuning  with  Synergistic Optimization for Relation Extraction}

\renewcommand{\shorttitle}{KnowPrompt}

\author{Xiang Chen, Ningyu Zhang}
\authornotemark[1]
\affiliation{%
  \institution{Zhejiang University \\ AZFT Joint Lab for Knowledge Engine \\ Hangzhou Innovation Center}
  \city{Hangzhou}
  \state{Zhejiang}
  \country{China}
}
\email{{xiang_chen,zhangningyu}@zju.edu.cn}

\author{Xin Xie, Shumin Deng}
\affiliation{%
  \institution{Zhejiang University \\ AZFT Joint Lab for Knowledge Engine \\ Hangzhou Innovation Center}
  \city{Hangzhou}
  \state{Zhejiang}
  \country{China}
}
\email{{xx2020,231sm}@zju.edu.cn}

\author{Yunzhi Yao}
\affiliation{%
  \institution{Zhejiang University \\ AZFT Joint Lab for Knowledge Engine \\ Hangzhou Innovation Center}
  \city{Hangzhou}
  \state{Zhejiang}
  \country{China}
}
\email{yyztodd@zju.edu.cn}

\author{Chuanqi Tan, Fei Huang}
\affiliation{%
  \institution{Alibaba Group}
  \city{Hangzhou}
  \state{Zhejiang}
  \country{China}
}
\email{{chuanqi.tcq,f.huang}@alibaba-inc.com}

\author{Luo Si}
\affiliation{%
  \institution{Alibaba Group}
  \city{Hangzhou}
  \state{Zhejiang}
  \country{China}
}
\email{luo.si@alibaba-inc.com}

\author{Huajun Chen}
\authornote{Corresponding author.}
\affiliation{%
  \institution{Zhejiang University \\ AZFT Joint Lab for Knowledge Engine \\ Hangzhou Innovation Center}
  \city{Hangzhou}
  \state{Zhejiang}
  \country{China}
}
\email{huajunsir@zju.edu.cn}

\renewcommand{\shortauthors}{Xiang Chen et al.}
\newcommand\tf[1]{\textbf{#1}}
\newcommand{\ours}{KnowPrompt}
\newcommand\ti[1]{\textit{#1}}
\newcommand\ts[1]{\textsc{#1}}
\newcommand\ttt[1]{\texttt{#1}}
\newcommand\mf[1]{\mathbf{#1}}
\newcommand{\dtrain}{\mathcal{D}_{\text{train}}}
\newcommand{\ddev}{\mathcal{D}_{\text{dev}}}
\newcommand{\dtest}{\mathcal{D}_{\text{test}}}
\newcommand{\seedset}{\mathcal{S}_{\text{seed}}}

\newcommand{\lsent}{\texttt{<}S_1\ttt{>}}
\newcommand{\lfirstsent}{\texttt{<}S_1\ttt{>}}
\newcommand{\lsecondsent}{\texttt{<}S_2\ttt{>}}
\newcommand{\sent}{\ttt{<}$S_1$\ttt{>}}
\newcommand{\firstsent}{\ttt{<}$S_1$\ttt{>}}
\newcommand{\secondsent}{\ttt{<}$S_2$\ttt{>}}
\newcommand{\xinput}{{X}_{\mathrm{in}}}
\newcommand{\xinputhat}{\tilde{X}_{\mathrm{in}}}
\newcommand{\xprompt}{{X}_{\mathrm{prompt}}}

\newcommand{\gen}{\mathrm{g}}

\newcommand{\template}{\mathcal{T}}
\newcommand{\lwordset}{\mathcal{W}}
\newcommand{\vocabulary}{\mathcal{V}}
\newcommand{\vocabularyhat}{\mathcal{V'}}
\newcommand{\labelset}{\mathcal{Y}}
\newcommand{\mapping}{\mathcal{M}}
\newcommand{\totalk}{K_{\text{tot}}}
\newcommand{\lm}{\mathcal{L}}
\newcommand{\cls}{\texttt{[CLS]}}
\newcommand{\sep}{\texttt{[SEP]}}
\newcommand{\mask}{\texttt{[MASK]}}
\newcommand{\tableindent}{~~}
\newcommand{\customcomment}[3]{\textcolor{#1}{[#2:#3]}}
\newcommand{\todo}[1]{\customcomment{red}{TODO}{#1}}
\newcommand{\chj}[1]{\customcomment{cyan}{CHJ}{#1}}


\begin{abstract}
  Recently, prompt-tuning has achieved promising results for specific few-shot classification tasks. The core idea of prompt-tuning is to insert text pieces (i.e., templates) into the input and transform a classification task into a masked language modeling problem. However, for relation extraction, determining an appropriate prompt template requires domain expertise, and it is cumbersome and time-consuming to obtain a suitable label word. Furthermore, there exists abundant semantic and prior knowledge among the relation labels that cannot be ignored. To this end,  we focus on incorporating knowledge among relation labels into prompt-tuning for relation extraction and propose a \textbf{Know}ledge-aware \textbf{Prompt}-tuning approach with synergistic optimization (\textbf{\ours}). Specifically, we inject latent knowledge contained in relation labels into prompt construction with learnable virtual type words and answer words. Then, we synergistically optimize their representation with structured constraints. Extensive experimental results on five datasets with standard and low-resource settings demonstrate the effectiveness of our approach. Our code and datasets are available in GitHub\footnote{
 \url{https://github.com/zjunlp/KnowPrompt}} for reproducibility.

\end{abstract}

\ccsdesc[500]{Information systems~Information extraction}

\keywords{Relation Extraction, Prompt-tuning,  Knowledge-aware}

\maketitle

\section{Introduction}
\label{intro}
Relation Extraction (RE) aims to extract structured knowledge from unstructured text and plays a critical role in information extraction and knowledge base construction. 
RE appeals to many researchers \cite{DBLP:conf/emnlp/ZhangZCAM17,DBLP:conf/emnlp/ZhangDSCZC18,alignkg,DBLP:conf/naacl/ZhangDSWCZC19,www20-tom,www21-shen,DBLP:journals/corr/abs-2101-01926} due to the capability to extract textual information and benefit many web applications, e.g., information retrieval, web mining, and question answering. 

\begin{figure}[t!]
    \centering
    \includegraphics[width=0.45\textwidth]{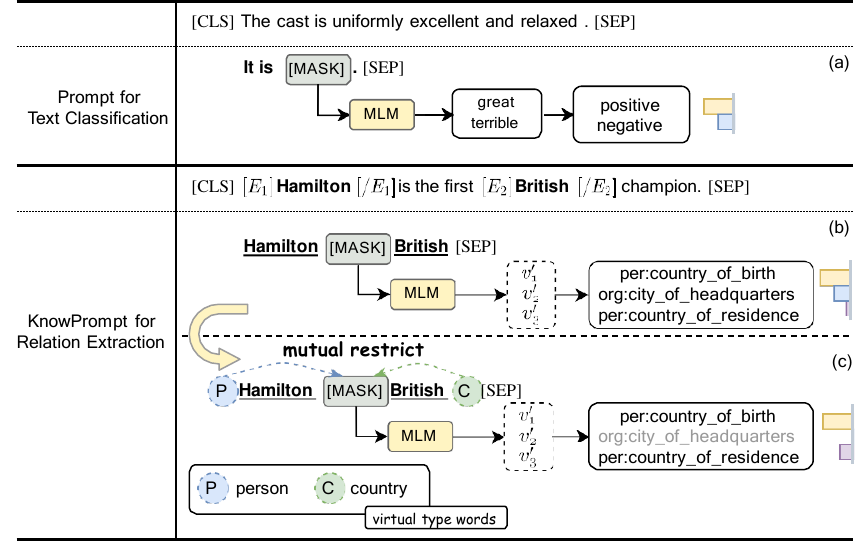}
    \caption{Examples of prompt-tuning to stimulate the knowledge of PLMs by formalizing specific tasks as cloze-style tasks. The $\mathbf{P}$ and $\mathbf{C}$  in  dashed balls represents the virtual type words with semantics close to person and country.}
    \label{fig:example}
\end{figure}

Previous self-supervised pre-trained language models (PLMs) such as BERT \cite{DBLP:conf/naacl/DevlinCLT19} have achieved state-of-the-art (SOTA) results in lots of RE benchmarks. 
However, since fine-tuning requires adding extra classifiers on top of PLMs and further training the models under classification objectives, their performance heavily depends on time-consuming and labor-intensive annotated data, making it hard to generalize well.
Recently, a series of studies using prompt-tuning \cite{DBLP:conf/eacl/SchickS21,DBLP:journals/corr/abs-2009-07118,ppt,typeprompt,knowledgeprompt} to address this issue: adopting the pre-trained LM directly as a predictor by completing a cloze task to bridge the gap between pre-training and fine-tuning. 
Prompt-tuning fuses the original input with the prompt template to predict $\texttt{[MASK]}$ and then maps the predicted label words to the corresponding class sets, which has induced better performances for PLMs on few-shot tasks.
As shown in Figure~\ref{fig:example} (a), a typical prompt for text classification consists of a template (e.g. ``<$S_1$> It is $\texttt{[MASK]}$ '') and a set of label words (``great'', ``terrible''etc.) as candidates to predict $\texttt{[MASK]}$. 
PLMs predict (``great'', ``terrible'', etc.) at the masked position to determine the label of the sentence ``<$S_1$>''. 
In a nutshell, prompt-tuning involves template engineering and verbalizer engineering, which aims to search for the best template and an answer space~\cite{liu2021pre}.

Despite the success of prompt-tuning PLMs for text classification tasks, there are still several non-trivial challenges for RE with prompt-tuning as follows:
on the one hand, determining the appropriate prompt template for RE requires domain expertise, and auto-constructing a high-performing prompt with input entities often requires additional computation cost for generation and verification \cite{schick2020automatically,DBLP:conf/eacl/SchickS21,shin2020eliciting,DBLP:journals/corr/abs-2012-15723};
on the other hand, the computational complexity of the label word search process is very high (e.g., usually exponentially depending on the number of categories) when the length of the relation label varies, and it is non-trivial to obtain a suitable target label word in the vocabulary to represent the specific relation label.
For example, the relation labels of $per:country\_of\_birth$ and $org:city\_of\_headquarters$ cannot specify a single suitable label word in the vocabulary.
In addition, there exists rich semantic knowledge among relation labels and structural knowledge implications among relational triples, which cannot be ignored.
For example, as shown in Figure \ref{fig:example} (b) and (c), if a pair of entities contains the semantics of ``person'' and ``country'', the prediction probability of the $\texttt{[MASK]}$ on the relation ``org:city\_of\_headquarters'' will be lower.
Conversely, the relation also restricts the types of its
subject and object entity.
Previous studies \cite{li2019improving,distiawan2019neural,bastos2021recon} indicate that incorporating the relational knowledge will provide evidence for RE.

To address those issues, we take the first step to inject knowledge into learnable prompts and propose a novel \textbf{Know}ledge-aware \textbf{Prompt}-tuning with synergistic optimization (\textbf{KnowPrompt}) approach for RE.  
We construct prompt with knowledge injection via learnable virtual answer words and virtual type words to alleviate labor-intensive prompt engineering (\S \ref{subsec:cons}).  
To be specific, instead of a regular verbalizer that mapping from one label word in vocabulary to the particular class,
we creatively propose to \textbf{leverage learnable virtual answer words by injecting in semantic knowledge to represent relation labels}.
Furthermore, we assign learnable virtual type words surrounding entities to hold the role of weakened \textsc{Type Marker}~\cite{DBLP:journals/corr/abs-2102-01373}, which are initialized with prior knowledge maintained in relation labels. 
Notably, we innovatively \textbf{leverage learnable virtual type words to dynamically adjust according to context rather than utilizing annotation of the entity type}, which may not be available in datasets.
Since there exist implicit structural constraints among entities and relations, and virtual words should be consistent with the surrounding contexts, we introduce synergistic optimization to obtain optimized virtual type and answer words (\S \ref{sec:opt}). 
Concretely, we propose a context-aware prompt calibration method with implicit structural constraints to inject structural knowledge implications among relational triples and associate prompt embeddings with each other. 

\section{Related Work}
\subsection{Relation Extraction}
Relation Extraction (RE) involves extracting the relation between two given entities based on their related context, which plays an essential task in information extraction and knowledge base construction.
Early approaches involve pattern-based methods~\citep{DBLP:conf/ijcai/Huffman95,DBLP:conf/aaai/CaliffM99}, CNN/RNN-based~\citep{DBLP:conf/emnlp/ZengLC015,DBLP:conf/acl/ZhouSTQLHX16, DBLP:conf/emnlp/ZhangZCAM17} and graph-based methods ~\cite{DBLP:conf/emnlp/Zhang0M18,DBLP:conf/acl/GuoZL19,guo2020learning}. 
With the recent advances in pre-trained language models~\citep{DBLP:conf/naacl/DevlinCLT19}, applying PLMs as the backbone of RE systems~\cite{www20-Zhou,wang2020finding,li2020logic,AliCG,PRGC,generative_triple,docunet} has become standard procedure. 
Several studies have shown that BERT-based models significantly outperform both RNN and graph-based models \cite{DBLP:conf/cikm/WuH19a,joshi2020spanbert,DBLP:conf/acl/YuSCY20}. 
Meanwhile, a series of knowledge-enhanced PLMs have been further explored, which use knowledge bases as additional information to
enhance PLMs. Among them, MTB\cite{baldini-soares-etal-2019-matching} propose matching the blanks based on BERT, which is a RE-oriented pre-trained method to learn relational patterns from text. 
SPANBERT~\cite{joshi2020spanbert} adopt knowledge to enhance learning objectives, 
KNOWBERT~\cite{peters2019knowledge} propose to incorporate knowledge into input features,
and LUKE~\cite{yamada2020luke} leverage knowledge to improve model architectures.
We compare with this line of work here for their promotion comes from relational knowledge of external sources. In contrast to them, we focus on learning from the text itself in the paper.
Recently, \citet{Fuzhao2021gdpnet} propose a multi-view graph based on BERT, achieving SOTA performance both on TACRED-Revisit~\cite{alt2020tacred} and DialogRE~\cite{DBLP:conf/acl/YuSCY20}.
Thus, we also choose the latest graph methods based on BERT for RE as our baselines to demonstrate the effectiveness of our \ours.

Some previous studies ~\cite{DBLP:journals/corr/abs-2112-10006} have focused on the few-shot setting since available annotated instances may be limited in practice. ~\citet{DBLP:conf/emnlp/HanZYWYLS18,DBLP:conf/aaai/GaoH0S19,DBLP:conf/aaai/GaoHX0LLS20,DBLP:conf/icml/QuGXT20,bridge_triple,DBLP:conf/coling/DongYXGHLLLS20} propose approaches for few-shot RE based on meta-learning or metric learning, with the aim of developing models that can be trained with only a few labeled sentences and nonetheless generalize well. 
In contrast to previous N-way K-shot approaches, \citet{DBLP:journals/corr/abs-2012-15723} utilize a  setting that is relatively practical both for acquiring a few annotations (e.g., 16 examples per class) and efficiently training.


\subsection{Prompt-tuning}
Prompt-tuning methods are fueled by the birth of GPT-3~\cite{DBLP:conf/nips/BrownMRSKDNSSAA20} and have achieved outstanding performance in widespread NLP tasks.
With appropriate manual prompts, series of studies \cite{liu2021pre,ben2021pada,lester2021power,DBLP:journals/corr/abs-2103-08493,reynolds2021prompt,lu2021fantastically} have been proposed, demonstrating the advancement of prompt-tuning. 
\citet{hu2021knowledgeable} propose to incorporate external knowledge into the verbalizer with calibration.  
\citet{ding2021prompt} apply prompt-tuning to entity typing with prompt-learning by constructing an entity-oriented verbalizer and templates. 
To avoid labor-intensive prompt design, automatic searches for discrete prompts have been extensively explored.  
\citet{schick2020automatically,DBLP:journals/corr/abs-2012-15723} first explore the automatic generation of ans words and templates.
\citet{shin2020eliciting} further propose gradient-guided search to generate the template and label word in vocabulary automatically.  
Recently, some continuous prompts have also been proposed~\cite{li2021prefix,warp,liu2021gpt,paras}, which focus on utilizing learnable continuous embeddings as prompt templates rather than label words.
However, these works can not adapted to RE directly.

\begin{figure*}[!htbp] 
\centering 
\includegraphics[width=0.79\textwidth]{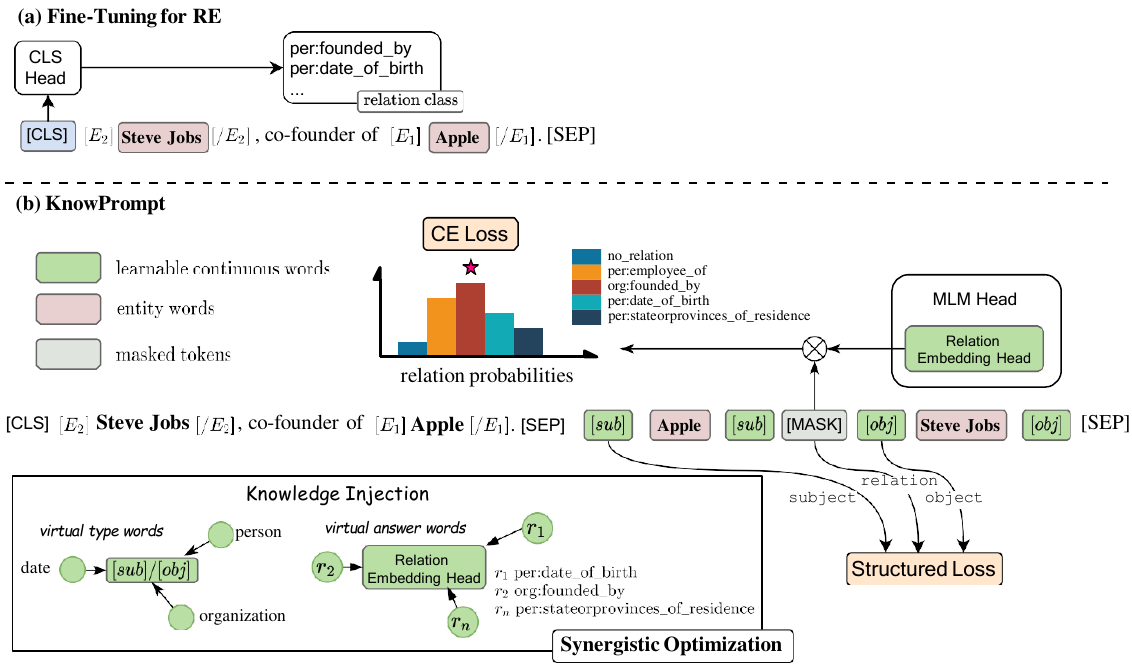} 
\caption{Model architecture of fine-tuning for RE (Figure a), and proposed KnowPrompt (Figure b) approach (Best viewed in color). The answer word described in the paper refers to the virtual answer word we proposed.} 
\label{fig:model}
\end{figure*}

For relation extraction, \citet{ptr}  proposes a model called PTR, which creatively applies logic rules to construct prompts with several sub-prompts. 
Compared with their approach, our approach has three significant differences. 
\textbf{Firstly}, we \textbf{propose virtual answer words to represent specific relation labels rather than multiple sub-prompt} in PTR. Essentially, our method is \textbf{model-agnostic} that can be applied to generative LMs, while PTR fails due to its sub-prompt mechanism.
\textbf{Secondly}, we construct prompt with knowledge injection via learnable virtual type words and virtual answer words to alleviate labor-intensive prompt engineering rather than predefined rules; thus, our method is more \textbf{flexible} and can generalize to different RE datasets easily.
\textbf{Thirdly}, we synergistically optimize virtual type words and answer words with knowledge constraints and associate prompt embeddings with each other. 

\section{Background}
\label{sec:setup}
An RE dataset can be denoted as $\mathcal{D} = \{\mathcal{X}, \mathcal{Y}\}$, where $\mathcal{X}$ is the set of examples and $\mathcal{Y}$ is the set of relation labels. 
For each example $x = \{w_1, w_2, w_s \ldots w_o,\ldots  w_n \}$, the goal of RE is to predict the relation $y \in \mathcal{Y}$ between subject entity $w_s$ and object entity $w_o$ (since one entity may have multiple tokens, we simply utilize $w_s$ and $w_o$ to represent all entities briefly.).

\subsection{Fine-tuning PLMs for RE} 

Given a pre-trained language model (PLM) $\mathcal{L}$ for RE, previous fine-tuning methods first convert the instance $x=\{w_1, w_s \ldots  w_o,\ldots  w_n \}$ into an input sequence of PLM, such as $\texttt{[CLS]} x \texttt{[SEP]}$. 
The PLM $\mathcal{L}$ encodes the input sequence into the corresponding output hidden vectors such as $\mathbf{h} =\{\mathbf{h}_{\texttt{[CLS]}}, \mathbf{h}_1, \mathbf{h}_s, \ldots, \mathbf{h}_o,\ldots, \mathbf{h}_{\texttt{[SEP]}}\}$. 
Normally, a $\texttt{[CLS]}$ head is utilized to compute the probability distribution over the class set $\mathcal{Y}$ with the softmax function $p(\cdot|x) = \texttt{Softmax}(\mathbf{W}\mathbf{h}_{\texttt{[CLS]}})$, where $\mathbf{h}_{\texttt{[CLS]}}$ is the output embedding of $\texttt{[CLS]}$ and $\mathbf{W}$ is a randomly initialized matrix that needs to be optimized. 
The parameters of $\mathcal{L}$ and $\mathbf{W}$ are fine-tuned by minimizing the cross-entropy loss over $p(y|x)$ on the entire $\mathcal{X}$.

\subsection{Prompt-Tuning of PLMs}
Prompt-tuning is proposed to bridge the gap between the pre-training tasks and downstream tasks.
The challenge is to construct an appropriate template $\template(\cdot)$ and label words $\mathcal{V}$, which are collectively referred to as a prompt $\mathcal{P}$. 
For each instance $x$, the template is leveraged to map $x$ to prompt the input $x_\text{prompt} = T(x)$. 
Concretely, template $\template(\cdot)$ involves the location and number of added additional words.
$\vocabulary$ refers to a set of label words in the vocabulary of a language model $\lm$, and $\mapping \colon \labelset \rightarrow \vocabulary$ is an injective mapping that connects task labels to label words $\vocabulary$. 
In addition to retaining the original tokens in $x$, one or more $\texttt{[MASK]}$ is placed into $x_\text{prompt}$ for $\mathcal{L}$ to fill the label words. 
As $\mathcal{L}$ can predict the right token at the masked position, we can formalize $p(y|x)$ with the probability distribution over $\mathcal{V}$ at the masked position, that is, $p(y|x) = p([\texttt{MASK}] = \mapping(y) | x_\text{prompt})$.
Taking the binary sentiment classification task described as an example, we set the template $T(\cdot) = \text{``} \cdot \text{It is} \texttt{[MASK]} \text{.''}$ and map $x$ to $x_\text{prompt} = \text{``} x \text{ It is} \texttt{[MASK]} \text{.''}$. 
We can then obtain the hidden vector of $\texttt{[MASK]}$ by encoding  $x_\text{prompt}$ by $\mathcal{L}$ and produce a probability distribution $p(\texttt{[MASK]}|x_\text{prompt})$, describing which tokens of $\mathcal{V}$ are suitable for replacing the \texttt{[MASK]} word. 
Since previous study for prompt-learning involves searching or generating label words
here, we simply set  $\mapping(y=\text{``}positive\text{''}) \rightarrow \text{``}great\text{''}$ and $\mapping(y=\text{``}negative\text{''}) \rightarrow \text{``}terrible\text{''}$ as examples. According to whether $\mathcal{L}$ predicts "great" or "terrible", we can identify if the label of instance $x$ is either $positive$ or $negative$.

\section{Methodology}
In this section, we  introduce our \textbf{Know}ledge-aware \textbf{Prompt}-tuning with synergistic optimization ({\ours}) approach to be aware of semantic and prior knowledge contained in relation labels for relation extraction. As shown in Figure \ref{fig:model},
we elucidate the details of how to construct (\S \ref{subsec:cons}), optimize (\S \ref{sec:opt}) the {\ours}.

\subsection{Prompt Construction with Knowledge Injection}
\label{subsec:cons}
Because a typical prompt consists of two parts, namely a template and a set of label words, we propose the construction of virtual type words and virtual answer words with knowledge injection for the RE task.

\textbf{Entity Knowledge Injection.}
Note that \textsc{Type Marker}~\cite{DBLP:journals/corr/abs-2102-01373} methods can additionally introduce the type information of entities to improve performance but require additional annotation of type information, which is not always available in datasets.
However, we can obtain the \textbf{scope of the potential entity types} with prior knowledge contained in a specific relation, rather than annotation.
For instance, given the relation ``per:country\_of\_birth'', it is evident that the subject entity matching this relation belongs to ``person'' and the object entity matching this relation belongs to ``country''. 
Intuitively, we estimate the \textbf{prior distributions} $\phi_{sub}$ and $\phi_{obj}$ over the candidate set $C_{sub}$ and $C_{obj}$ of  potential entity types, respectively, according to the relation class, where the prior distributions are estimated by frequency statistics. Take $C_{sub}$ and $C_{obj}$ of partial relation labels listed in the Table~\ref{tab:temp} as an example, the prior distributions for $C_{sub}$ can be counted as:  $\phi_{sub}= {\{\text{``}organization\text{''}: 3/6,\text{``}person\text{''}: 3/6 \}}$.
Because of this, we assign virtual type words around the entities, which are initialized with aggregated embeddings of the set of potential entity types. 
Since initialized virtual type words are not precise types for specific entities, those learnable virtual type words can dynamically adjust according to context and play the \textbf{weakened role of \textsc{Type Marker}}  for RE. 
The specific initialization method is as follows:
\begin{equation}
\begin{aligned}
 \mf{\hat{e}}_{[sub]} &= \sum \mf{\phi}_{sub} \cdot \mf{e} \left( I \left( \mf{C}_{sub} \right) \right),
\end{aligned}
\end{equation}

\begin{equation}
\begin{aligned}
\mf{\hat{e}}_{[obj]} &= \sum \mf{\phi}_{obj} \cdot \mf{e} \left( I \left( \mf{C}_{obj} \right) \right),
\end{aligned}
\end{equation}

where $\mf{\hat{e}}_{[sub]}$ and $\mf{\hat{e}}_{[obj]}$ represent the embeddings of virtual type words surrounding the subject and object entities, $I \left(  \cdot \right)$ is the deduplication operations on sets,  and $\mf{e}$ is the word-embedding layer of $\mathcal{L}$.
Since the virtual type words designed based on the prior knowledge within relation labels can initially perceive the range of entity types, it can be further optimized according to context to express semantic information close to the actual entity type, holding the role similar to \textsc{Typer Marker}.

\begin{table*}[t]
\center
\small
\caption{Examples of some relations of the datasets TACREV, and relation-specific $\mf{C}_{sub}$ 
,$\mf{C}_{obj}$
and $\mf{C}_{r}$.}
\label{tab:temp}
\scalebox{0.92}{
\begin{tabular}{l|c|c|c}
\toprule
Relation Labels        
& $\mf{C}_{sub}$ 
& $\mf{C}_{obj}$
& $\mf{C}_{r}$ (Disassembled Relation Prepared for Virtual Answer Words)  \\
\midrule
per:country\_of\_birth & person 
& country
& \text{\{``country'', ``of'', ``birth'' \}}\\
per:data\_of\_death & person 
& data 
& \text{\{``data'', ``of'', ``death'' \}} \\
per:schools\_attended
& person & organization 
& \text{\{``school'',``attended'\}}  \\
org:alternate\_names
& organization 
& organization 
& \text{\{``alternate'', ``names'' \}} \\
org:city\_of\_headquarters & organization & city 
& \text{\{``city'',  ``of'', ``headquarters'' \}} \\
org:number\_of\_employees/members 
& organization & number  
& \text{\{``number'',  ``of'', ``employees'', ``members'' \}} \\
\bottomrule
\end{tabular}}
\centering
\end{table*} 

\textbf{Relation Knowledge Injection.}
Previous studies on prompt-tuning usually form a one-one mapping between one label word in the vocabulary and one task label by automatic generation, which maintains large computational complexity of the search process and fails to leverage the abundant semantic knowledge in relation labels for RE.
To this end, we assume that there exists a  virtual answer word $v' \in \mathcal{V'}$  in the vocabulary space of PLMs, which can represent the implicit semantics of the relation.
From this perspective, we expand the \texttt{MLM Head} layer of $\mathcal{L}$  with extra learnable relation embeddings as the virtual answer word sets $\mathcal{V'}$ to completely represent the corresponding relation labels $\mathcal{Y}$.
Thus, we can reformalize $p(y|x)$ with the probability distribution over $\mathcal{V'}$ at the masked position.
We propose to \textbf{encodes
semantic knowledge about the label and facilitates the process of RE}.
Concretely, we set the $\phi_{R}=[\phi_{r1}, \phi_{r2}, ... ,\phi_{rm}]$ and $\mf{C}_{R}=[\mf{C}_{r1}, \mf{C}_{r2}, ... ,\mf{C}_{rm}]$, where $\phi_{r}$ represent the probability distribution over the candidate set $\mf{C}_{r}$  of the semantic words of relation  by \textbf{disassembling} the relation label $r$, $m$ is the number of relation labels.
Furthermore, we adopt the weighted average function for $\phi_{r}$ to average embeddings of each words among $\mf{C}_{r}$ to initialize these relation embeddings, which can inject the semantic knowledge of relations.
The specific decomposition process is shown in Table~\ref{tab:temp}, and the learnable relation embedding of virtual answer word $v'=\mapping(y)$ is initialized as follows:
\begin{equation}
\begin{aligned} 
\mf{\hat{e}}_{[rel]}(v')=\phi_{r} \cdot \mf{e} \left( \mf{C}_{r}  \right),\\
\end{aligned}
\end{equation}
where $\mf{\hat{e}}_{[rel]}(v')$is the embedding of virtual label word $v'$,
$\mf{e}$ represents the word-embedding layer of $\mathcal{L}$.
It is noticed that the knowledgeable initialization process of virtual answer words may be regarded as a great anchor; we can further optimize them based on context to express optimal semantic information, leading to better performance.

\subsection{Synergistic Optimization with Knowledge Constraints}
\label{sec:opt}
Since there exist close interaction and connection between entity types and relation labels, and those virtual type words as well as answer words should be associated with the surrounding context, we further introduce a \textbf{synergistic optimization} method with implicit structural constraints over the \textbf{\textit{ parameter set}} $\{ \mf{\hat{e}}_{[sub]}, \mf{\hat{e}}_{[obj]}, \mf{\hat{e}}_{[rel]}(\mathcal{V'}) \}$ of virtual type words and virtual answer words.

\textbf{Context-aware Prompt Calibration.} 
Although our virtual type and answer words are initialized based on knowledge, they may not be optimal in the latent variable space. They should be associated with the surrounding context. 
Thus, further optimization is necessary by perceiving the context to calibrate their representation. 
Given the probability distribution  $p(y|x) = p([\texttt{MASK}] = \mathcal{V'} | x_\text{prompt})$ over $\mathcal{V'}$ at the masked position, we optimize the virtual type words as well as answer words by the loss function  computed as the cross-entropy between $\mathbf{y}$ and  $p(y|x)$ as follows:


\begin{equation}
\begin{aligned}
&\mathcal{J}_{\texttt{[MASK]}}=-\frac{1}{|\mathcal{X}|} \sum_{x\in \mathcal{X}} \mathbf{y} \log{p(y|x)} ,\\
\end{aligned}
\end{equation}

where $|\mathcal{X}|$ represents the numbers of the training dataset.
The learnable words may adaptively obtain optimal representations for prompt-tuning through a synergistic type and answer optimization.

\textbf{Implicit Structured  Constraints.}
To integrate structural knowledge into {\ours}, we adopt additional structured constraints to optimize prompts.
Specifically, we use a triplet $(s,r,o)$ to describe a relational fact; here, $s,o$ represent the virtual types of subject and object entities, respectively, and $r$ is the relation label within a pre-defined set of answer words $\mathcal{V'}$.
In {\ours}, instead of using pre-trained knowledge graph embeddings\footnote{Note that pre-trained knowledge graph embeddings are heterogeneous compared with pre-trained language model embeddings.}, we directly leverage the output embedding of virtual type words and virtual answer words through LMs to participate in the calculation.
We define the loss $\mathcal{J}_{\texttt{struct}}$ of implicit structured constraints  as follows:
\begin{equation} \label{equ:ke}
\begin{aligned}
    \mathcal{J}_{\texttt{structured}}=-\log{\sigma (\gamma-d_r(\mathbf{s}, \mathbf{o}))}\\
      -\sum_{i=1}^{n}\frac{1}{n}\log{\sigma(d_r(\mathbf{s_i^{\prime}}, \mathbf{o_i^{\prime}})-\gamma)},
\end{aligned}
\end{equation}

\begin{equation}\label{equ:transe}
    d_r(\mathbf{s}, \mathbf{o})=\|\mathbf{s}+\mathbf{r}-\mathbf{o}\|_2,
\end{equation}
 
where $(s_i^{\prime}, r, o_i^{\prime})$ are negative samples, $\gamma$ is the margin, $\sigma$ refers to the $\text{sigmoid}$ function and $d_r$ is the scoring function. 
For negative sampling, we assign the correct virtual answer words at the position of  {\texttt{[MASK]}} and randomly sample the subject entity or object entity and replace it with an irrelevant one to construct corrupt triples, in which the entity has an impossible type for the current relation.

\begin{table}[htbp]
\small
\center
\caption{
Statistics for RE datasets used in the paper, including numbers of relations and instances in the different split.
For dialogue-level DialogRE, instance refers to the number of documents.}
\label{tab:data_analysis} 
\scalebox{0.92}{
\begin{tabular}{l|r|r|r|c}
\toprule
Dataset & \textbf{\#~Train.} & \textbf{\#~Val.} & \textbf{\#~Test.} & \textbf{\#~Rel.} \\
\midrule
SemEval &6,507 &1,493 &2,717 &19\\
DialogRE & 5,963& 1,928& 1,858&36\\
TACRED &68,124 &22,631 &15,509 & 42\\
TACRED-Revisit &68,124 &22,631 &15,509 & 42\\
Re-TACRED &58,465 &19,584 &13,418 &40\\
\bottomrule
\end{tabular}
}
\end{table}

\subsection{Training Details}
Our approach has a two-stage optimization procedure. 
First, we synergistically optimize the parameter set $\{ \mf{\hat{e}}_{[sub]}, \mf{\hat{e}}_{[obj]}, \mf{\hat{e}}_{[rel]}(\mathcal{V'}) \}$ of virtual type words and virtual answer words with a large learning rate $lr_1$ to obtain the optimal prompt as follows:
\begin{equation}
    \mathcal{J}=\mathcal{J}_{\texttt{[MASK]}}+
    \lambda \mathcal{J}_\texttt{structured},
    \label{equ:loss}
\end{equation}
where $\lambda$ is the hyperparameter, and $\mathcal{J}_{\texttt{structured}}$ and $\mathcal{J}_{\texttt{[MASK]}}$ are the losses for the KE and \texttt{[MASK]} prediction, respectively. Second, based on the optimized virtual type words and answer words, we utilize the object function $\mathcal{J}_{\texttt{[MASK]}}$ to tune the parameters of the PLM with prompt (optimizing overall parameters) with a small learning rate $lr_2$.
For more experimental details, please refer to the Appendix.

\section{Experiments}

\subsection{Datasets}
For comprehensive experiments, we carry out our experiments on five RE datasets:
SemEval 2010 Task 8 (SemEval)~\cite{hendrickx2010semeval},
DialogRE~\cite{DBLP:conf/acl/YuSCY20},
TACRED~\cite{zhang2017position},
TACRED-Revisit~\cite{alt2020tacred}, Re-TACRED~\cite{stoica2021re}. 
Statistical details are provided in Table~\ref{tab:data_analysis}
and Appendix~\ref{app:data_analysis}:

\begin{table*}[!htb]
\center
\small
\caption{Standard RE performance of $F_1$ scores (\%) on different test sets. 
``w/o'' means that no additional data is used for pre-training and fine-tuning, yet ``w/'' means that the model uses extra data for tasks. It is worth noting that ``$\dagger$'' indicates we exceptionally rerun the code of {\ours} and PTR with \textsc{RoBERT\_base} for a fair comparison
with current SOTA models on DialogRE. 
Subscript in red represents advantages of {\ours} over the best results of baselines.
Best results are bold.}
\label{tab:supervised}
\scalebox{0.92}{
\begin{tabular}{l|c|c|c|c|c|c}
\hline
\toprule
\multicolumn{7}{c}{\textit{Standard Supervised Setting}}\\
\midrule
Methods   & Extra Data & SemEval  & DialogRE$\dagger$ & TACRED 
& TACRED-Revisit & Re-TACRED  \\
\midrule
\multicolumn{7}{c}{Fine-tuning pre-trained models}               \\
\midrule
\textsc{Fine-tuning}-[\textsc{Roberta}]~ & w/o & 87.6 &57.3 & 68.7 & 76.0 & 84.9     \\

\textsc{SpanBERT}~\cite{joshi2020spanbert}      
& w/ & - & - 
& 70.8 & 78.0  & 85.3        \\
\textsc{KnowBERT}~\cite{peters2019knowledge}
& w/  & 89.1  & -  
& 71.5 & 79.3  & 89.1     \\
\textsc{LUKE}~\cite{yamada2020luke} & w/  & -   & - 
& \textbf{72.7} & 80.6  & -         \\
\textsc{MTB}~\cite{baldini-soares-etal-2019-matching}  &  w/ & 89.5   & -  & 70.1 & - & -      \\
\textsc{GDPNet} ~\cite{Fuzhao2021gdpnet}
& w/o & -  & 64.9 
& 71.5 & 79.3  & -       \\
\textsc{Dual} ~\cite{semdialogre}
& w/o   & -  & 67.3 & - & -  & -      \\

\midrule
\multicolumn{7}{c}{Prompt-tuning pre-trained models}               \\

\midrule
\textsc{PTR}-[\textsc{Roberta}]~\cite{ptr}  & w/o & 89.9  & 63.2 & 72.4 & 81.4   & 90.9     \\
\textbf{\textsc{\ours}}-[\textsc{Roberta}] & w/o  
& \textbf{90.2} 
\small{\color{red}{(+0.3)}} 
& \textbf{68.6}  
\small{\color{red}{(+5.4)}} 
& 72.4  
\small{\color{blue}{(-0.3)}} 
& \textbf{82.4}  
\small{\color{red}{(+1.0)}}  
& \textbf{91.3}  
\small{\color{red}{(+0.4)}}  
    \\
\bottomrule
\hline
\end{tabular}
}

\end{table*}

\subsection{Experimental Settings}
For fine-tuning vanilla PLMs and our {\ours}, we utilize \textsc{RoBERT\_large} for all experiments to make a fair comparison (except for DialogRE, we adopt \textsc{RoBERTa\_base} to compare with previous methods). 
For test metrics, we use micro $F_1$ scores of RE as the primary metric to evaluate models, considering that $F_1$ scores can assess the overall performance of precision and recall. We use different settings for standard and low-resource experiments.All detailed settings for our {\ours}, Fine-tuning and PTR can be found in the Appendix ~\ref{app:imp1}, ~\ref{app:imp2} and ~\ref{app:imp3}.

\textbf{Standard Setting.}
In the standard setting, we utilize full $\dtrain$ to fine-tune. 
Considering that entity information is essential for models to understand relational semantics, a series of knowledge-enhanced PLMs have been further explored using knowledge graphs as additional information to enhance PLMs. 
Specifically, we select \textsc{SpanBERT}~\cite{joshi2020spanbert}, \textsc{KnowBERT}~\cite{peters2019knowledge}, \textsc{LUKE}~\cite{yamada2020luke}, and \textsc{MTB}~\cite{baldini-soares-etal-2019-matching} as our strong baselines, which are typical models that use external knowledge to enhance learning objectives, input features, model architectures, or pre-training strategies. 
We also compare several SOTA models on DialogRE, in which one challenge is that each entity pair has more than one relation.

\textbf{Low-Resource Setting.}
We conducted 8-, 16-, and 32-shot experiments following LM-BFF~\cite{DBLP:journals/corr/abs-2012-15723,ptr} to measure the average performance across five different randomly sampled data based on every experiment using a fixed set of seeds $\seedset$. 
Specifically, we sample $k$ instances of each class from the initial training and validation sets to form the few-shot training and validation sets.

\subsection{Main Results}

\textbf{Standard Result.}
As shown in Table~\ref{tab:supervised}, the knowledge-enhanced PLMs yield better performance than the vanilla \textsc{Fine-tuning}.
This result illustrates that it is practical to inject task-specific knowledge to enhance models, indicating that simply fine-tuning PLMs cannot perceive knowledge obtained from pre-training. 

Note that our {\ours} achieves improvements over all baselines and even achieves better performance than those knowledge-enhanced models,  which use knowledge as data augmentation or architecture enhancement during fine-tuning.
On the other hand,
even if the task-specific knowledge is already contained in knowledge-enhanced PLMs such as 
\textsc{LUKE}, \textsc{KnowBERT} \textsc{SpanBERT} and \textsc{MTB}
, it is difficult for fine-tuning to stimulate the knowledge for downstream tasks.
Overall, we believe that the development of prompt-tuning is imperative and 
{\ours} is a simple and effective prompt-tuning paradigm for RE.

\textbf{Low-Resource Result.}
From Table \ref{tab:few_shot},  {\ours} appears to be more beneficial in low-resource settings. 
We find that {\ours} consistently outperforms the baseline method \textsc{Fine-tuning}, \textsc{GDPNet}, and  \textsc{PTR} in all datasets, especially in the 8-shot and 16-shot experiments. 
Specifically, our model can obtain gains of up to \textbf{22.4\%} and \textbf{13.2\%}  absolute improvement on average compared with \textsc{fine-tuning}.
As $K$ increases from $8$ to $32$, the improvement in our {\ours} over the other three methods decreases gradually.
For 32-shot, we think that the number of labeled instances is sufficient. 
Thus, those rich semantic knowledge injected in our approach may induce fewer gains. 
We also observe that \textsc{GDPNet} even performs worse than \textsc{Fine-tuning} for 8-shot, which reveals that the complex SOTA model in the standard supervised setting may fall off the altar when the data are extremely scarce. 

\begin{table*}[ht]
\centering
\small
\caption{\label{tab:movie}Low-resource RE performance of $F_1$ scores (\%) on different test sets. We use $K = 8,16,32$ (\# examples per class) for few-shot experiments.
Subscript in red represents the advantages of {\ours} over the results of  \textsc{Fine-tuning}.}
\label{tab:few_shot}
\scalebox{0.92}{
\begin{tabular}{l|l|c|c|c|c|c|c}
\hline
\toprule
\multicolumn{8}{c}{\textit{Low-Resource Setting}}\\
\midrule
    Split & Methods  & SemEval  & DialogRE$\dagger$  & TACRED
    & TACRED-Revisit  & Re-TACRED  & Average \\
\midrule
    \multirow{4}{*}{K=8} 
    & \textsc{Fine-tuning}  & 41.3 & 29.8 & 12.2
    & 13.5 & 28.5  & 25.1 \\
    & \textsc{GDPNet}  & 42.0  & 28.6 & 11.8 & 12.3 & 29.0  & 24.7 \\
    & \textsc{PTR} & 70.5 & 35.5 & 28.1 &28.7  & 51.5  & 42.9 \\
    & \textbf{\textsc{\ours}} 
    & \tf{74.3} \small{\color{red}{(+33.0)}}
    & \tf{43.8} \small{\color{red}{(+14.0)}}
    & \tf{32.0} \small{\color{red}{(+19.8)}}
    & \tf{32.1} \small{\color{red}{(+18.6)}}
    & \tf{55.3} \small{\color{red}{(+26.8)}}
    & \tf{47.5} \small{\color{red}{(+22.4)}} \\
\midrule

    \multirow{4}{*}{K=16} 
    & \textsc{Fine-tuning}  & 65.2  & 40.8 & 21.5
    & 22.3 & 49.5  & 39.9 \\
    & \textsc{GDPNet}  & 67.5  & 42.5 & 22.5
    & 23.8 & 50.0   & 41.3 \\
    & \textsc{PTR}  & 81.3 & 43.5 & 30.7
    &31.4 & 56.2  & 48.6 \\
    & \textbf{\textsc{\ours}}
    & \tf{82.9}  \small{\color{red}{(+17.7)}}
    & \tf{50.8}  \small{\color{red}{(+10.0)}}
    & \tf{35.4} \small{\color{red}{(+13.9)}}
    & \tf{33.1} \small{\color{red}{(+10.8)}}
    & \tf{63.3} \small{\color{red}{(+13.8)}}
    & \tf{53.1} \small{\color{red}{(+13.2)}} \\

\midrule
    \multirow{4}{*}{K=32} 
    & \textsc{Fine-tuning}  & 80.1 & 49.7 & 28.0
    & 28.2 & 56.0  & 48.4 \\
    & \textsc{GDPNet}  & 81.2  & 50.2 & 28.8
    & 29.1 & 56.5   & 49.2 \\
    & \textsc{PTR}  & 84.2 & 49.5 & 32.1
    & 32.4 & 62.1  & 52.1 \\
    & \textbf{\textsc{\ours}} 
    & \tf{84.8} \small{\color{red}{(+4.7)}} 
    & \tf{55.3} \small{\color{red}{(+3.6)}} 
    & \tf{36.5} \small{\color{red}{(+8.5)}} 
     & \tf{34.7} \small{\color{red}{(+6.5)}}  
    & \tf{65.0} \small{\color{red}{(+9.0)}} 
    & \tf{55.3} \small{\color{red}{(+6.9)}} \\

\bottomrule
\hline
\end{tabular}
}

\end{table*}

\begin{table}[htbp]
    \begin{center}
    \small
    \caption{Ablation study  on SemEval, VAW and VTW refers to virtual answer words and type words.
    }
    \label{tab:pair_ablation}
    \scalebox{0.9}{
    \centering
    \resizebox{1.0\columnwidth}{!}{%
    \begin{tabular}{l c c c c}
    \toprule
    \tf{Method}
    & \tf{K=8} & \tf{K=16} & \tf{K=32} & \tf{Full} \\
    \toprule
    \textsc{\ours}    & \tf{74.3}   &\tf{82.9}  & \tf{84.8}      &\tf{90.2}   \\
    \midrule
    \tableindent -VAW   
    & 68.2  &  72.7
    & 75.9  & 85.2   \\
    \tableindent -Knowledge Injection for VAW
    & 52.5  &  78.0
    &  80.2 & 88.0   \\
    \tableindent -VTW  
    &72.8   & 80.3
    & 82.9  & 88.7 \\
    \tableindent -Knowledge Injection for VTW
    & 68.8  &  79.5
    & 81.6  & 88.5   \\
    \tableindent -Structured Constrains  
    &73.5  & 81.2      
    & 83.6 & 89.3 \\

    \bottomrule
    \end{tabular}
    }
    }
    \end{center}
\end{table}

\textbf{Comparison between {\ours} and Prompt Tuning Methods.}
The typical prompt-tuning methods perform outstandingly on text classification tasks (e.g., sentiment analysis and NLI), such as LM-BFF, but they don't involve RE application. Thus we cannot rerun their code for RE tasks.
To our best knowledge, PTR is the only method that uses prompts for RE, which is a wonderful job and works in the same period as our \ours. 
Thus, we make a comprehensively comparative analysis between {\ours} and PTR, and summarize the comparison in Table ~\ref{tab:comparison}. The specific analysis is as follows:

\newcommand{\tabincell}[2]{\begin{tabular}{@{}#1@{}}#2\end{tabular}}
\begin{table*}[!htbp]
\small
\center
\caption{Interpreting representation of \textbf{virtual type words}.  
We obtain the hidden state  $\mathbf{h}_\texttt{[sub]}, \mathbf{h}_\texttt{[obj]}$ through the PLM,  then adopt \texttt{MLM Head} over them to explore which words in the vocabulary is nearest the virtual type words. 
}
\label{tab:marker}
\resizebox{0.98\textwidth}{!}{
\begin{tabular}{l|p{.15\textwidth}|p{.15\textwidth}}
\toprule
Input Example of our \ours   & Top 3 words around \texttt{[sub]}  & Top 3 words around \texttt{[obj]}     \\
\midrule
 \tabincell{l}{
 $\textbf{x}$:\texttt{[CLS]} It sold $[E_1]$ \textbf{ALICO} $[/E_1]$ to $[E_2]$ \textbf{MetLife Inc}  $[E_2]$ for \$ 162 billion. \texttt{[SEP]}  \\
 \texttt{[sub]} \textbf{ALICO} \texttt{[sub]} \texttt{[MASK]} \texttt{[obj]} \textbf{MetLife Inc}
 \texttt{[obj]}. \texttt{[SEP]} \\
 $\textbf{y}$: "$org:member\_of$"
 }
 &    \tabincell{l}{organization\\group\\corporation}        &  \tabincell{l}{company\\plc\\organization}              \\ 
 \midrule
  \tabincell{l}{
  $\textbf{x}$: \texttt{[CLS]}  $[E_1]$  \textbf{Ismael Rukwago} $[/E_1]$, a senior $[E_2]$ \textbf{ADF} $[E_2]$ commander, denied any involvement. \texttt{[SEP]}  \\
 \texttt{[sub]}\textbf{Ismael Rukwago}  \texttt{[sub]} \texttt{[MASK]} \texttt{[obj]}  \textbf{ADF}
 \texttt{[obj]}. \texttt{[SEP]}   \\
 $\textbf{y}$: "$per:employee\_of$" }
 &    \tabincell{l}{person\\commander\\colonel}        &  \tabincell{l}{intelligence\\organization\\command}              \\
\bottomrule
\end{tabular}
}

\end{table*}

\textbf{Firstly}, PTR adopt a fixed number of multi-token answer form and LM-BFF leverage actual label word with single-token answer form, while {\ours} propose \textbf{virtual answer word with single-token answer form}. Thus,  PTR needs to manually formulate rules, which is more labor-intensive. LM-BFF requires expensive label search due to its search process exponentially depending on the number of categories.

\textbf{Secondly}, essentially attributed to to the difference of answer form, our {\ours} and LM-BFF is \textbf{model-agnostic}  and can be plugged into different kinds of PLMs (As show in Figure~\ref{fig:differentLMs}, our method can adopted on GPT-2), while PTR fails to generalize to generative LMs due to it's nultiple discontinuous $\texttt{[MASK]}$ prediction.

\textbf{Thirdly}, above experiments, demonstrates that {\ours} is comprehensively comparable to the PTR, and can perform better in low-resource scenarios.
Especially for DialogRE, a multi-label classification task, our method exceeded PTR by approximately 5.4 points in the standard supervised settings. It may be attributed to the rule method used by PTR that forcing multiple mask predictions will confuse multi-label predictions.

In a nutshell, the above analysis proves that {\ours} is more flexible and widely applicable; meanwhile, it can be aware of knowledge and stimulate it to serve downstream tasks better. 

\begin{table}[htbp]
\small
\center
\caption{
Comparative statistics between {\ours} and PTR, including (1)Answer Form of prompt; (2) labor-intensive; (3) MA refers to whether model-agnostic ; (4) ML refers to the ability of multi-label learning;  and (4) CC refers to the computational complexity.}
\label{tab:comparison} 
\scalebox{0.92}{
\begin{tabular}{l|c|c|c|c|c}
\toprule
\textbf{Method} & \textbf{\#~Answer Form}  & \textbf{\#~Labor} 
& \textbf{\#~MA} 
& \textbf{\#~ML}
& \textbf{\#~CC}\\
\midrule
LM-BFF & single-token  & normal & yes & - & high \\
PTR & multi-token  & normal & no & normal & norm \\
{Ours} & single-token  & small & yes & better &norm \\
\bottomrule
\end{tabular}
}
\end{table}

\subsection{Ablation Study  on \ours}
\noindent\textbf{Effect of Virtual Answer Words Modules:} To prove the effects of the virtual answer words and its knowledge injection, we conduct the ablation study, and the results are shown in Table \ref{tab:pair_ablation}.
For \textit{-VAW}, we adopt one specific token in the relation label as the label word without optimization, and for \textit{-Knowledge Injection for VAW}, we 
randomly initialize the virtual answer words to conduct optimization. 
Specifically, removing the knowledge injection for virtual answer words has the most significant effect, causing the relation F1 score to drop from 74.3\% to 52.5\% in the 8-shot setting. It also reveals that the injection of semantic knowledge maintained in relation labels is critical for relation extraction, especially in few-shot scenarios.

\noindent\textbf{Effect of Virtual Type Words Modules:} We also conduct an ablation study to validate the effectiveness of the design of virtual type words.
As for \textit{-VTW}, we directly remove virtual type words, and for \textit{-Knowledge Injection for VTW}, we 
randomly initialize the virtual type words to conduct optimization. 
In the 8-shot setting, the performance of the directly removing virtual type words drops from 74.3 to 72.8, while randomly initialized virtual type words decrease the performance to 68.1, which is much lower than 72.8.
This phenomenon may be related to the  
noise disturbance caused by random initialization, while as the instance increase, the impact of knowledge injection gradually diminishes.
Despite this, it still demonstrates that our design of knowledge injection for virtual type words is effective for relation extraction.

\noindent\textbf{Effect of Structured Constrains:}
Moreover, \textit{-Structured Constraints} refer to the model without implicit structural constraints, which indicates no direct correlations between entities and relations. The result demonstrates that structured constraints certainly improve model performance, probably, because they can force the virtual answer words and type words to interact with each other better.

Overall, the result reveals that all modules contribute to the final performance.
We further notice that virtual answer words with knowledge injection are more sensitive to performance and highly beneficial for \ours, especially in low-resource settings.

\section{Analysis and Discussion}

\subsection{Can {\ours} Applied to Other LMs?}
\begin{figure}[!htbp] 
\centering 
\small
\includegraphics[width=0.32\textwidth]{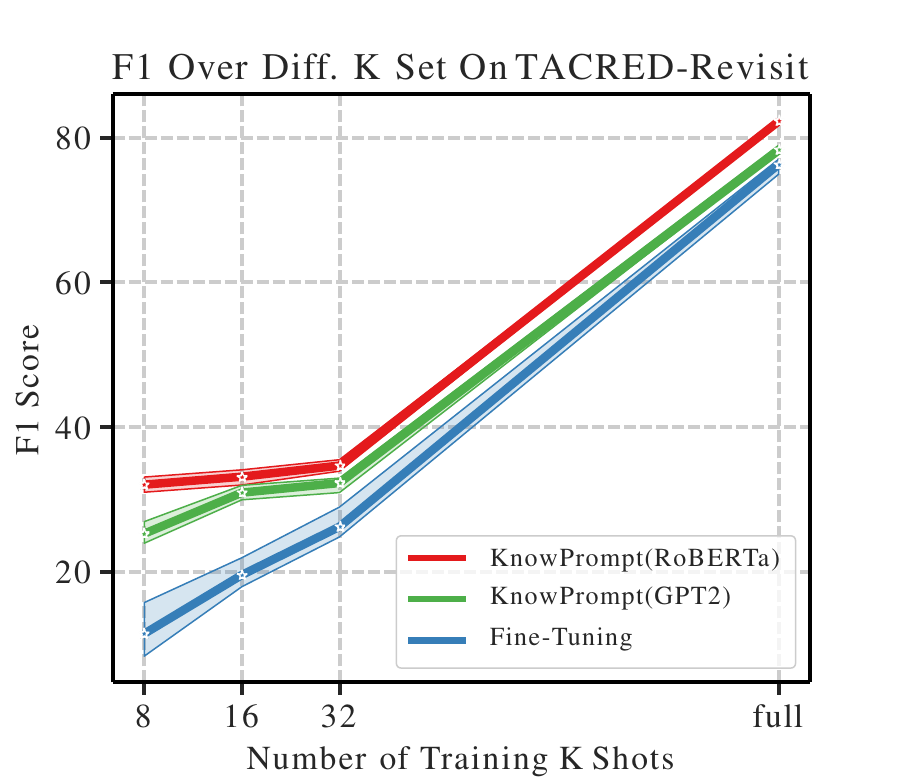} 
\caption{RoBERT-large vs. GPT-2 results on TACRED-Revisit dataset regarding different K (instances per class). 
} 
\label{fig:differentLMs} 
\end{figure}

Since we focus on MLM (e.g., RoBERTa) in the main experiments, we further extend our {\ours} to autoregressive LMs like GPT-2. 
Specifically, we directly append the prompt template with [MASK] at the end of the input sequence for GPT-2. We further apply the relation embedding head by extending the word embedding layer in PLMs; thus, GPT2 can generate virtual answer words.
We first notice that fine-tuning leads to poor performance with high variance in the low-resource setting, while {\ours} based on RoBERTa or GPT-2 can achieve impressive improvement with low variance compared with  \textsc{Fine-tuning}.
As shown in Figure \ref{fig:differentLMs},  {\ours} based on GPT-2 obtains the results on par of the model with RoBERTa-large, which reveals our method can unearth the potential of GPT-2 to make it perform well in natural language understanding tasks such as RE. 
This finding also indicates that our method is model-agnostic and can be plugged into different kinds of PLMs.  

\subsection{Interpreting Representation Space of Virtual Answer Words}
\begin{figure}[!htbp] 
\centering 
\includegraphics[width=0.38\textwidth]{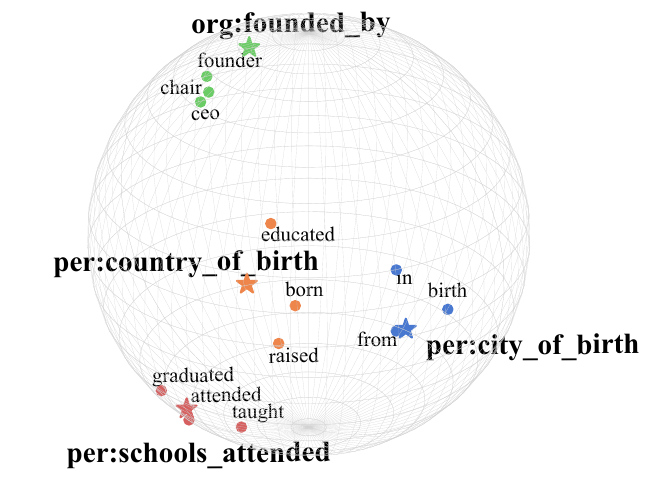} 
\caption{A 3D visualization of several  relation representations (\textbf{virtual answer words}) optimized in {\ours} on TACRED-Revisit dataset using t-SNE and normalization. } 
\label{fig:scatter} 
\end{figure}

Since the embeddings of virtual answer words $\{ \mf{\hat{e}}_{[rel]}(\mathcal{V'}) \}$ are initialized with \textbf{semantic knowledge of relation type itself}, and further learned in continuous space, it is intuitive to explore  what precisely the optimized virtual answer word is. 
We use t-SNE and normalization to map the embedding to 3 dimension space and make a 3D visualization of several sampled virtual answer words in the TACRED-Revisit dataset. We also get the top3 tokens nearest the virtual answer word by calculating the $L_2$ distance between the embedding of the virtual answer word and the actual word in the vocabulary.
For example, ``$org:founded\_by$''  referred to as green  $\color{green}{\star}$ in Figure \ref{fig:scatter} represents the relation type, which is learned by optimizing virtual answer words in vocabulary space, and the ``$founder$'',``$chair$'' and ``$ceo$'' referred to as green $\color{green}{\bullet}$ are the words closest to the it.
It reveals that virtual answer words learned in vocabulary space are semantic and intuitive.
To some extent, our proposed virtual answer words are similar to prototypical representation for relation labels. 
This inspired us to further expand knowprompt into the field of prototype representation learning in the future, which can also be applied to other NLP tasks with prompt-tuning.

\subsection{Interpreting Representation Space of virtual type words}
Since we initialize the virtual type words with the average embedding of candidate types of head and tail entities through \textbf{prior knowledge maintained in the relation labels}, and synergistically optimize them ($\{ \mf{\hat{e}}_{[sub]}, \mf{\hat{e}}_{[obj]} \}$) with virtual answer words based on context. 
To this end, we further conduct further analysis to investigate that what semantics do the optimized type words express and whether virtual type words can adaptively reflect the entity types based on context as shown in Table~\ref{tab:marker}.
Specifically, we apply the MLM head over the position of the virtual type words to get the output representation and get the top-3 words in vocabulary nearest the virtual type words according to the $L_2$ distance of embeddings between virtual type words and other words. 
We observe that thanks to the synergistic optimization with knowledge constraints, those learned virtual type words can dynamically adjust according to context and play a reminder role for RE.

\section{Conclusion and Future Work}
In this paper, we present {\ours} for relation extraction, which mainly includes knowledge-aware prompt construction and synergistic optimization with knowledge constraints. 
In the future, we plan to explore two directions, including:
(i) extending to semi-supervised setting to further leverage unlabelled data;
(ii) extending to lifelong learning, whereas prompt should be optimized with adaptive tasks.

\begin{acks}
This work is funded by NSFC91846204/NSFCU19B2027, National Key R\&D Program of China (Funding No.SQ2018YFC000004), Zhejiang Provincial Natural Science Foundation of China (No. LGG22F030011), Ningbo Natural Science Foundation (2021J190), and Yongjiang Talent Introduction Programme (2021A-156-G). Our work is supported by Information Technology Center and State Key Lab of CAD\&CG, ZheJiang University.
\end{acks}


\bibliographystyle{ACM-Reference-Format}
\balance
\bibliography{sample-base}


\appendix

\section{Detailed Statistics of Dataset}
\label{app:data_analysis}
For comprehensive experiments, we carry out our experiments on five relaction extraction datasets: TACRED~\cite{zhang2017position},
TACREV~\cite{alt2020tacred}, Re-TACRED~\cite{stoica2021re}, SemEval 2010 Task 8 (SemEval)~\cite{hendrickx2010semeval}, and DialogRE~\cite{DBLP:conf/acl/YuSCY20}.
A brief introduction to these data is as follows:

\textbf{TACRED:} one large-scale sentence-level relation extraction dataset drawn from the yearly TACKBP4 challenge, which contains more than 106K sentences. It involves 42 different relations (41 common relation types and a special “no relation” type). The subject mentions in TACRED are person and organization, while object mentions are in 16 fine-grained types, including date, number, etc.

\textbf{TACRED-Revisit:} one dataset built based on the original TACRED dataset. They find out and correct the errors in the original development set and test set of TACRED, while the training set was left intact. 

\textbf{Re-TACRED:} another version of TACRED dataset. They address some shortcomings of the original TACRED dataset, refactor its training set, development set and test set. Re-TACRED also modifies a few relation types, finally resulting in a dataset with $40$ relation types.

\textbf{SemEval:} a traditional dataset in relation classification containing $10,717$ annotated examples covering $9$ relations with two directions and one special relation ``no\_relation''.



\textbf{DialogRE:} DialogRE is the first human-annotated dialogue-level RE dataset. It contains 1,788 dialogues originating from the complete transcripts of a famous American television situation comedy. It is multi-label classification, as each entity pair may posses more than one relation.

\section{Implementation Details for \ours}
\label{app:imp1}

This section details the training procedures and hyperparameters for each of the datasets. We utilize Pytorch to conduct experiments with 8 Nvidia 3090 GPUs. All optimizations are performed with the AdamW optimizer with a linear warmup of learning rate over the first 10\% of gradient updates to a maximum value, then linear decay over the remainder of the training. Gradients are clipped if their norm exceeded 1.0, margin $\gamma$, $\lambda$ and weight decay on all non-bias parameters are set to 1, 0.001 and 0.01.
A grid search is used for hyperparameter tuning (maximum values bolded below).

\subsection{Standard Supervised Setting}

The hyper-parameter search space is shown as follows:
\begin{itemize}
\item learning rate $lr_1$ of synergistic optimization for virtual template and anchor words.
[5e-5,\textbf{1e-4}, 2e-4 ]
\item learning rate $lr_2$ of optimization for overall parameters.
[1e-5, 2e-5, \textbf{3e-5}, 5e-5]
\item number epochs 5 (for dialogre as 20)
\item batch size: 16  (for tacrev, retacred and dialogre  as 8)
\item max seq length: 256  (for tacrev, retacred and dialogre as 512)
\item gradient accumulation steps: 1 (for dialogre as 4)
\end{itemize}

\subsection{Low-Resource Setting}

The hyper-parameter search space is shown as follows:
\begin{itemize}
\item learning rate $lr_1$ of synergistic optimization for virtual template and anchor words:
[5e-5,\textbf{1e-4}, 2e-4 ]
\item learning rate $lr_2$ of optimization for overall parameters:
[1e-5, \textbf{2e-5}, 3e-5, 5e-5]
\item number of epochs: 30
\item batch size: 16  (for tacrev, retacred and dialogre  as 8)
\item max seq length: 256  (for tacrev, retacred and dialogre as 512)
\item gradient accumulation steps: 1 (for dialogre as 4)
\end{itemize}

\section{Implementation Details for Fine-tuning}
\label{app:imp2}

The fine-tuning method is conducted as shown in Figure~\ref{fig:model}, which is both equipped with the same entity marker in the raw text for a fair comparison.
The hyper-parameters such as batch size, epoch, and learning rate are the same as \ours.

\section{Implementation Details for PTR}
\label{app:imp3}

Since PTR does not conduct experiments on DialogRE in standard supervised setting and SemEval and DialogRE  in few-shot settings, 
we rerun its public code to supplement the experiments we described above with these data and scenarios.
As for SemEval, the experiment process completely follows the original setting in his code, while for DialogRE, we modify his code to more adapt to the setting of this data set. The specific hyper-parameters such as batch size, epoch, and learning rate are the same as \ours.

\end{document}